\documentclass[conference]{IEEEtran}
\usepackage{amsmath,amssymb,amsfonts}
\usepackage{algorithmic}
\usepackage{graphicx}
\usepackage{textcomp}
\usepackage{booktabs}
\usepackage{xcolor}
\usepackage{natbib}
\usepackage{cite}
\usepackage{adjustbox}
\usepackage{resizegather}
\usepackage{algorithm}
\usepackage{subcaption}
\usepackage{multirow}

\setcitestyle{authoryear,open={(},close={)}}

\def\BibTeX{{\rm B\kern-.05em{\sc i\kern-.025em b}\kern-.08em
    T\kern-.1667em\lower.7ex\hbox{E}\kern-.125emX}}
\usepackage{hyperref}

\begin{document}

\title{RRT and RRT* Using Vehicle Dynamics}

\author{\IEEEauthorblockN{Abhish Khanal}
\IEEEauthorblockA{
Department of Computer Science \\
George Mason University \\
Fairfax, Virginia, USA}
}

\maketitle

\begin{abstract}
The trajectory derived from RRT and RRT* is zagged. A holonomic drive is able to follow this trajectory. But real-life vehicle which has dynamical constraints cannot follow this trajectory. In this work, we are going to modify the RRT and RRT* algorithm to generate a trajectory that a vehicle with dynamical constraint can follow. The continuous nature of a steering control and acceleration control in a real-world vehicle introduces the complexity in its model. To introduce constraint in the vehicle's motion, while reducing the number of control and hence complexity, we are modeling our vehicle as a Dubins car. A Dubins car has only three controls (turning left, turning right, and moving forward) with a fixed velocity which makes our model simple. We use dubins curve (path that dubins car can follow) to trace the trajectory in RRT and RRT* algorithm.

\end{abstract}


\section{Introduction}
The main objective of planning is to find a feasible path from the start point to the goal point in a 2D space while avoiding collision with the environment. Planning can be done by using the local information about the surrounding that the robot is in. There are two ways in which planning can be done. Online planning: (Generating trajectory incrementally during the robot’s motion), and Offline Planning (Computing the entire path to the goal before motion from known map). Generally, online planning is done when there is no information about the robot's surrounding or the environment is unknown. This work focuses on the offline planning approach. The work done in this project uses RRT and RRT* algorithms for non-holonomic robots. The RRT algorithm \citep{LaValle1998RapidlyexploringRT} is based on the incremental construction of a search tree that attempts to rapidly and uniformly explore obstacle-free segments in the configuration space. After the search tree is successfully created, a simple search among the branches of the tree can result in a collision-free path between any two points in the robot’s environment. One of the disadvantages of RRT is that no matter how many points you sample, the path length doesn’t converge to the optimal path length. This shortcoming is improved in the RRT* algorithm. When the number of sample points approaches infinity, the RRT* algorithm delivers the shortest possible path to the goal \citep{optimality}. The basic principle of the RRT* algorithm is similar to RRT but RRT* updates the neighbor points if the point can be reached closer from the start location following the way of the newly sampled point.  
\\
The assumption, however, made for these approaches is that the vehicle is holonomic. A holonomic vehicle can move in any direction. However, real-life vehicles have constraints in their motion. When a vehicle is in a parking lot, it cannot just move sideways to the left or right side of the parking lot spot. This is because the vehicle has differential constraints. Therefore, the vehicle has to move forward and adjust its position to go to the left or right side of the parking lot. \\
In this work, we have focused on implementing RRT and RRT* algorithms with a vehicle that has dynamical constraints.
This is challenging because the steering and acceleration control in a real-world vehicle is continuous. Because of different steering angle and acceleration, there can be multiple path from point A to point B that the vehicle can take. In this project, a Dubin’s car is selected as a representative model of the car-like vehicle with differential constraints. A dubins car only has three controls: moving forward, turning left at a constant turning radius, turning right at a constant turning radius. Lester dubins proves \citep{10.2307/2372560} that there are only six paths that are shortest path from point A to point B. Hence our complexity is reduced. We use the shortest path from these six combinations in our RRT and RRT* algorithm.
\\
The report is structured as follows. In section II differential constraints are presented through a mathematical model of Dubins car. Section III describes how to create dubins path from point A to point B in a 2D space. Section IV describes modified RRT and RRT* algorithm based on dubins path. Section V presents simulation results. Finally, the work is concluded in section VI with discussion and conclusion.

\section{Dubins Car Model}
A holonomic drive is a drive that can move in any directions. In these drives, the controllable degree of freedom is equal to total degrees of freedom. A non-holonomic drive is a drive that has differential constraints, also known as velocity constraints, because of which the drive has fewer actions available than the degree of freedom. 
\\
\begin{figure}[h]
        \centering
        \includegraphics[width=6cm]{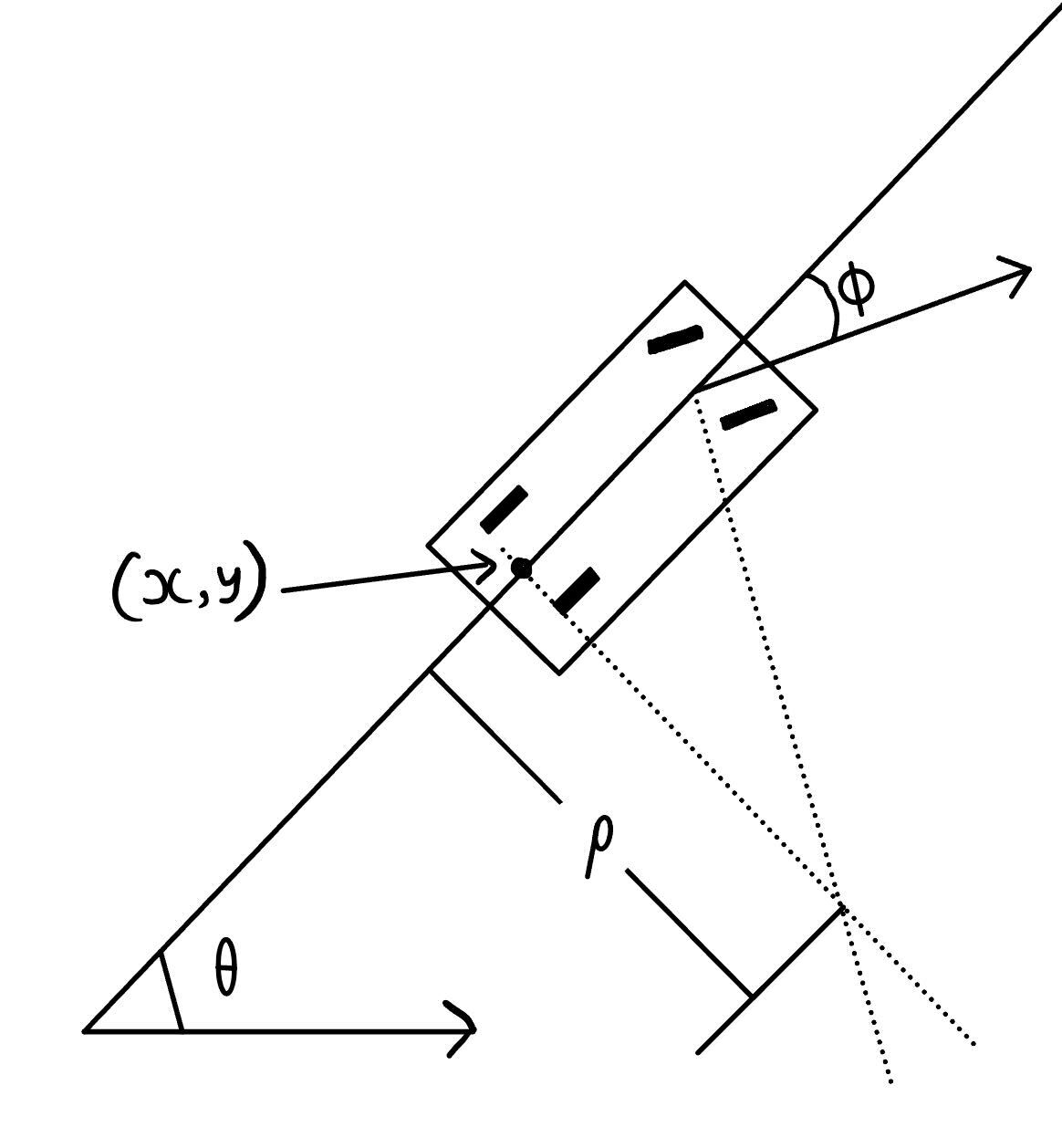}
        \caption{Dubins car model}
        \label{fig:dubinscar}
\end{figure}
The model of a Simple car is one of the examples of non-holonomic drive with differential constraints. If the speed of a Simple car is restricted to have positive values only (the car can only move forward), and can turn only at maximum steering angle, the model of Dubin’s car is obtained. This car-like model is shown in Fig \ref{fig:dubinscar}. Therefore, Dubin’s car can be defined as wheeled non-holonomic mobile robot that moves only forward with constant velocity and a maximum steering angle $\phi_{max}$, which corresponds with a minimum turning radius $\rho_{min}$. Based on \citep{10.5555/1213331} the vehicle shown in Fig. 1 can be considered as a rigid body that moves in the plane. In configuration space, the robot has three degrees of freedom denoted by the vector $q = (x, y, \theta)$. 

In a Dubin's car, we only have two motions. Move straight or turn (right or left) at a full steering angle. Since the speed s and the steering angle $\phi$ are only two variables that can be controlled, the control variables are defined using two-dimensional vector as follows: 
$$u = (u_s,u_{\phi})$$ 

The robot motion in 2D space is defined as, 

\begin{equation}
  \begin{aligned}
    \Dot{x} = u_scos(\theta) \\
    \Dot{y}= u_ssin(\theta) \\
    \Dot{\theta} = \frac{u_s}{L}tan(u_{\phi})
    \label{current_rel1}
  \end{aligned}
\end{equation}

Equation \ref{current_rel1} describes a simple Dubin’s car model. In order to complete this model, it is needed to specify allowed ranges for control variable.   
According to the assumption of a dubins car model, the dubins car should have a fixed forward velocity. 
$$u_s \in \{0,1\}$$
Considering the assumption that the vehicle can't rotate in-place, the steering angle should satisfy,
$$\phi_{max} < \frac{\pi}{2}, \phi \leq \phi_{max}$$

\section{Dubins path from point A to point B}

Unlike RRT and RRT* for holonomic robots, when we have non-holonomic drive, we also need to consider the heading that the vehicle is facing towards. So a vehicle is represented as a point $(x,y,\theta)$ where, x and y are the coordinates of the vehicle in a 2D space, and $\theta$ is the direction that the robot is facing towards. \\
Suppose we have start point $Ts (x_1,y_1,\theta_1)$ and end point $T_e(x_2,x_2,\theta_2)$. We have to find dubins path from $T_s$ to $T_e$. We have a dubins car model which can either turn right by turning steering fully towards right, turn left by turning steering fully towards left, or move straight. Hence we have three controls (turn left $(L)$, turn right $(R)$, straight $(S)$. If we have to go from point A to point B, we can reach there in shortest distance by following one of six trajectories from set $D$. where,

$$D = \{RSR,RSL,LSR,LSL,RLR,LRL\}$$ 

A work on classification of dubins set \citep{SHKEL2001179} finds the segment length for controls $(R, S, L)$ of the dubins path starting from origin to path length D along the x-axis for all of the six combinations of dubins curve.

Let $L_v$(for left turn), $R_v$(for right turn) and $S_v$(for straight motion) be three operator that transforms any arbitrary points $(x,y,\theta) \in \mathbb{R}^3$ into its corresponding image point in $\mathbb{R}^3$. Then,
\begin{gather*}
    L_v(x,y,\theta) = (x + sin(\theta + v) - sin(\theta), y - cos(\theta + v) + cos(\theta),\theta + v)
    \\
    R_v(x,y,\theta) = (x - sin(\theta - v) + sin(\theta), y + cos(\theta - v) - cos(\theta),\theta - v)
    \\
    S_v(x,y,\theta) = (x +  v*cos(\theta), y + v*sin(\theta),\theta)
\end{gather*}

\begin{figure}[h]
\centering
\begin{subfigure}{.25\textwidth}
  \centering
  \includegraphics[width=1.2\linewidth]{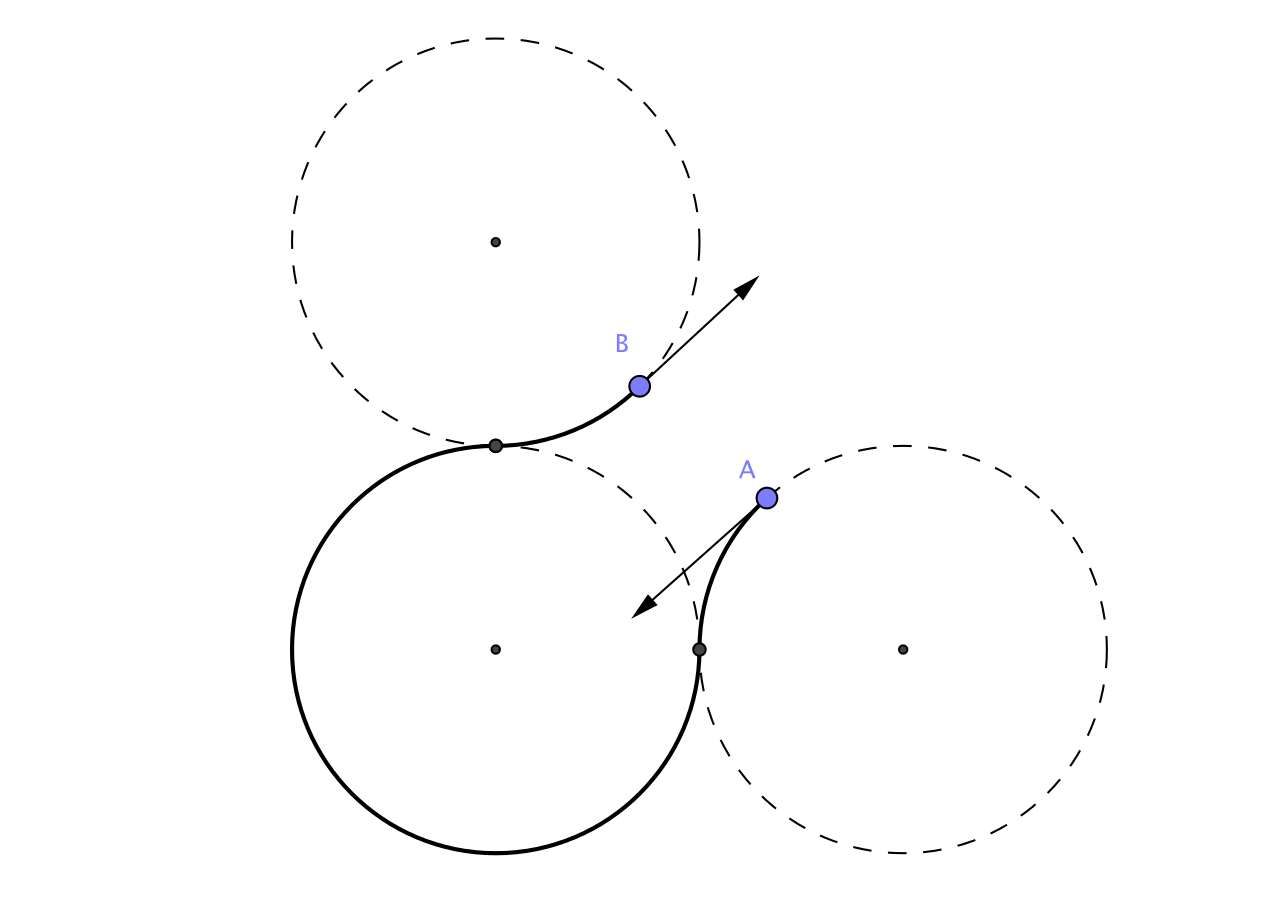}
  \caption{LRL Trajectory}
  \label{fig:lrl}
\end{subfigure}%
\begin{subfigure}{.25\textwidth}
  \centering
  \includegraphics[width=0.8\linewidth]{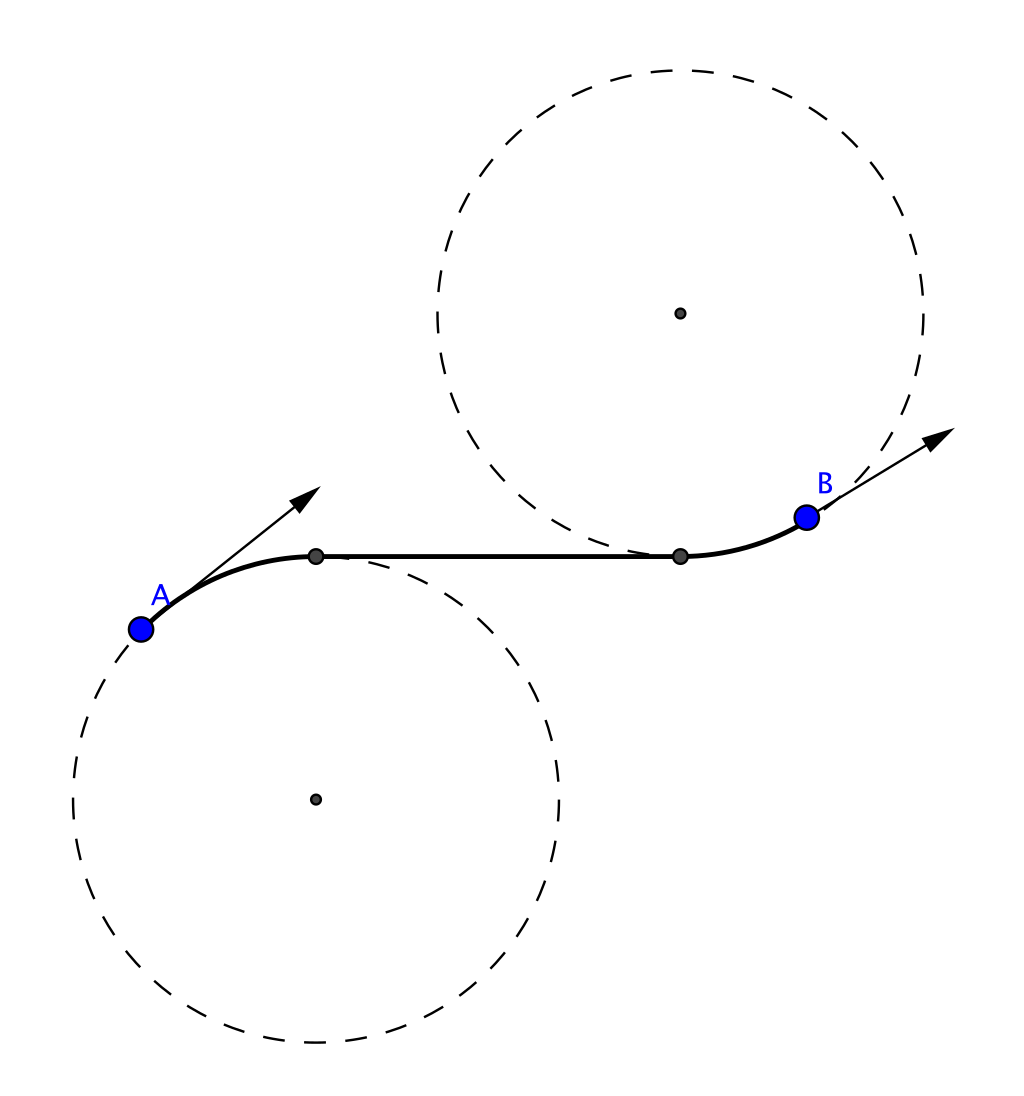}
  \caption{RSL Trajectory}
  \label{fig:rsl}
\end{subfigure}
\caption{LRL and RSL combination from dubins set of path}
\label{fig:dubins}
\end{figure}
where, index $v$ indicates that the motion has been along the $(R,L,S)$ segment of length $v$. Let $t$ be the path length of the first segment, $p$ be the path length of the second segment, and $q$ be the path length of third segment traced by dubins car after applying three control $(L,S,R)$. The dubins path length is then given by,
$$ L = t + p + q$$
Let the initial configuration of dubins car be $(0,0,\alpha)$ and final configuration be $(d,0,\beta)$. The path length of 3 segments in all the 6 combinations of dubins path $D = \{RSR,RSL,LSR,LSL,RLR,LRL\}$ are:
\begin{itemize}
    \item For $L_q (S_p (L_t (0, 0, \alpha))) = (d, 0, \beta)$:
    \begin{gather*}
    t_{lsl} = -\alpha + arctan \frac{cos\beta - cos\alpha}{d + sin\alpha - sin\beta}\{mod2\pi\} \\
    p_{lsl} = \sqrt{2 + d^2 - 2 cos (\alpha - \beta) + 2d(sin\alpha - sin\beta)} \\
    q_{lsl} = \beta - arctan \frac{cos\beta - cos\alpha}{d + sin\alpha - sin\beta}\{mod2\pi\}
    \end{gather*}
    The overall length for LSL trajectory is:
    $$L_{lsl} = -\alpha + \beta + p_{lsl}$$

    \item For $R_q (S_p (R_t (0, 0, \alpha))) = (d, 0, \beta)$:
    \begin{gather*}
    t_{lsr} = \alpha - arctan \frac{cos\beta - cos\alpha}{d - sin\alpha + sin\beta}\{mod2\pi\} \\
    p_{rsr} = \sqrt{2 + d^2 - 2 cos (\alpha - \beta) + 2d(sin\alpha - sin\beta)} \\
    q_{rsr} = -\beta\{mod2\pi\} + arctan \frac{cos\beta - cos\alpha}{d - sin\alpha + sin\beta}\{mod2\pi\}
    \end{gather*}
    The overall length for LSL trajectory is:
    $$L_{rsr} = \alpha - \beta + p_{rsr}$$

    \item For $R_q (S_p (L_t (0, 0, \alpha))) = (d, 0, \beta)$:
    \begin{gather*}
    t_{lsr} = \Bigg(-\alpha + arctan \frac{-cos\beta - cos\alpha}{d + sin\alpha + sin\beta} -arctan \bigg(\frac{-2}{p_{lsr}}\bigg)\Bigg)\{mod2\pi\} \\
    p_{lsr} = \sqrt{-2 + d^2 - 2 cos (\alpha - \beta) + 2d(sin\alpha + sin\beta)} \\
    q_{lsr} = \beta - arctan \frac{-cos\beta - cos\alpha}{d + sin\alpha + sin\beta} +arctan \bigg(\frac{-2}{p_{lsr}}\bigg)\{mod2\pi\}
    \end{gather*}
    The overall length for LSR trajectory is:
    $$L_{lsl} = \alpha - \beta + 2t_{lsr} + p_{lsr}$$
    
    \item For $L_q (S_p (R_t (0, 0, \alpha))) = (d, 0, \beta)$:
    \begin{gather*}
    t_{rsl} = \alpha - arctan \frac{cos\beta + cos\alpha}{d - sin\alpha - sin\beta} +arctan \bigg(\frac{2}{p_{rsl}}\bigg)\{mod2\pi\} \\
    p_{rsl} = \sqrt{-2 + d^2 + 2 cos (\alpha - \beta) - 2d(sin\alpha + sin\beta)} \\
    q_{rsl} = \beta\bigg(mod2\pi\bigg) - arctan \frac{cos\beta + cos\alpha}{d - sin\alpha - sin\beta} +arctan \bigg(\frac{2}{p_{rsl}}\bigg)\{mod2\pi\}
    \end{gather*}
    The overall length for RSL trajectory is:
    $$L_{rsl} = -\alpha + \beta + 2t_{rsl} + p_{rsl}$$
    
    \item For $R_q (L_p (R_t (0, 0, \alpha))) = (d, 0, \beta)$:
    \begin{gather*}
    t_{rlr} = \alpha - arctan \frac{-cos\beta + cos\alpha}{d - sin\alpha + sin\beta} + \bigg(\frac{p_{rlr}}{2}\bigg)\{mod2\pi\} \\
    p_{rlr} = arccos \frac{1}{8}\bigg(6 - d^2 + 2 cos(\alpha - \beta) + 2d (sin\alpha - sin\beta)\bigg) \\
    q_{rlr} = \alpha -\beta + t_{rlr} + p_{rlr} \{mod2\pi\}
    \end{gather*}
    The overall length for RSL trajectory is:
    $$L_{rlr} = \alpha - \beta + 2p_{rlr}$$
    
    \item For $L_q (R_p (L_t (0, 0, \alpha))) = (d, 0, \beta)$:
    \begin{gather*}
    t_{lrl} = \Bigg(-\alpha + arctan \frac{cos\beta - cos\alpha}{d + sin\alpha - sin\beta} + \bigg(\frac{p_{lrl}}{2}\bigg)\Bigg)\{mod2\pi\} \\
    p_{lrl} = arccos \frac{1}{8}\bigg(6 - d^2 + 2 cos(\alpha - \beta) + 2d (sin\alpha - sin\beta)\bigg)\{mod2\pi\} \\
    q_{lrl} = -\alpha + \beta\{mod2\pi\} + 2p_{lrl} \{mod2\pi\}
    \end{gather*}
    The overall length for RSL trajectory is:
    $$L_{lrl} = -\alpha + \beta + 2p_{lrl}$$
\end{itemize}

After we find the segment length, we can interpolate the points according to the control (Left, Right, or Straight) to get the trajectory. We take the combination of the path which has the minimum path length to get the dubins path.  
$$path = argmin\{L_{lrl},L_{rlr},L_{rsl},L_{lsl},L_{rsr},L_{lsl}\}$$
These equations of path length of segment $(t,p,q)$ requires start point as origin (0,0) and end point as (0,d), some distance 'd' from origin along x-axis. To find the trajectory for arbitrary start and end points in a 2D space, we use matrix transformation to transform the line joining arbitrary start point and end point as a line along x-axis with start point as origin. After we find the path length, we interpolate the points according to the controls $(L,S,R)$ and apply inverse transformation to get the trajectory.

\section{RRT and RRT* based on dubins path}
\begin{algorithm}[h]
\begin{algorithmic} 
\STATE N = 0
\STATE all\_links = [start\_position]
\WHILE{$N \neq n\_iter$}
\STATE $new\_point = find\_random\_point()$
\STATE \emph{find closest link to new\_point}
\STATE \emph{steer using \textbf{dubins path} towards new point}
\IF{(no dubins path) or (path collides)}
    \STATE do nothing
\ELSE
\STATE all\_links.append(new\_link)
\ENDIF
\STATE increment N
\ENDWHILE
\end{algorithmic}
\caption{Pseudocode of RRT with vehicle dynamics}
\label{algo_RRT}
\end{algorithm}

\begin{algorithm}[h]
\begin{algorithmic} 
\STATE N = 0
\STATE all\_links = [start\_position]
\WHILE{$N \neq n\_iter$}
\STATE $new\_point = find\_random\_point()$
\STATE \emph{find closest link to new\_point}
\STATE \emph{steer using \textbf{dubins path} towards new point}
\IF{(no dubins path) or (path collides)}
    \STATE do nothing
\ELSE
\STATE all\_links.append(new\_link)
\ENDIF

\STATE $neighbor = findNeighbor(new\_point)$
\IF{$neighbor.path\_length > new\_point.path\_length + dubins\_distance(neighbor,new\_point)$}
    \IF{new path doesn't collide with obstacle}
        \STATE reroute neighbor from new point
    \ELSE
        \STATE do nothing
    \ENDIF
\ENDIF
\STATE $N++$
\ENDWHILE
\end{algorithmic}
\caption{Pseudocode of RRT* with vehicle dynamics}
\label{algo_RRT*}
\end{algorithm}

The modified RRT and RRT* algorithm is shown in algorithm \ref{algo_RRT} and algorithm \ref{algo_RRT*} respectively.
The changes introduced here compared to the original algorithm for RRT  is that, the steer function is modified to include dubins path rather than a straight line path. Furthermore, in RRT* when neighbor is updated, the path is replaced with dubins path.

\section*{Simulation Results}
\begin{figure}[h]
\centering
\begin{subfigure}{.25\textwidth}
  \centering
  \includegraphics[width=\linewidth]{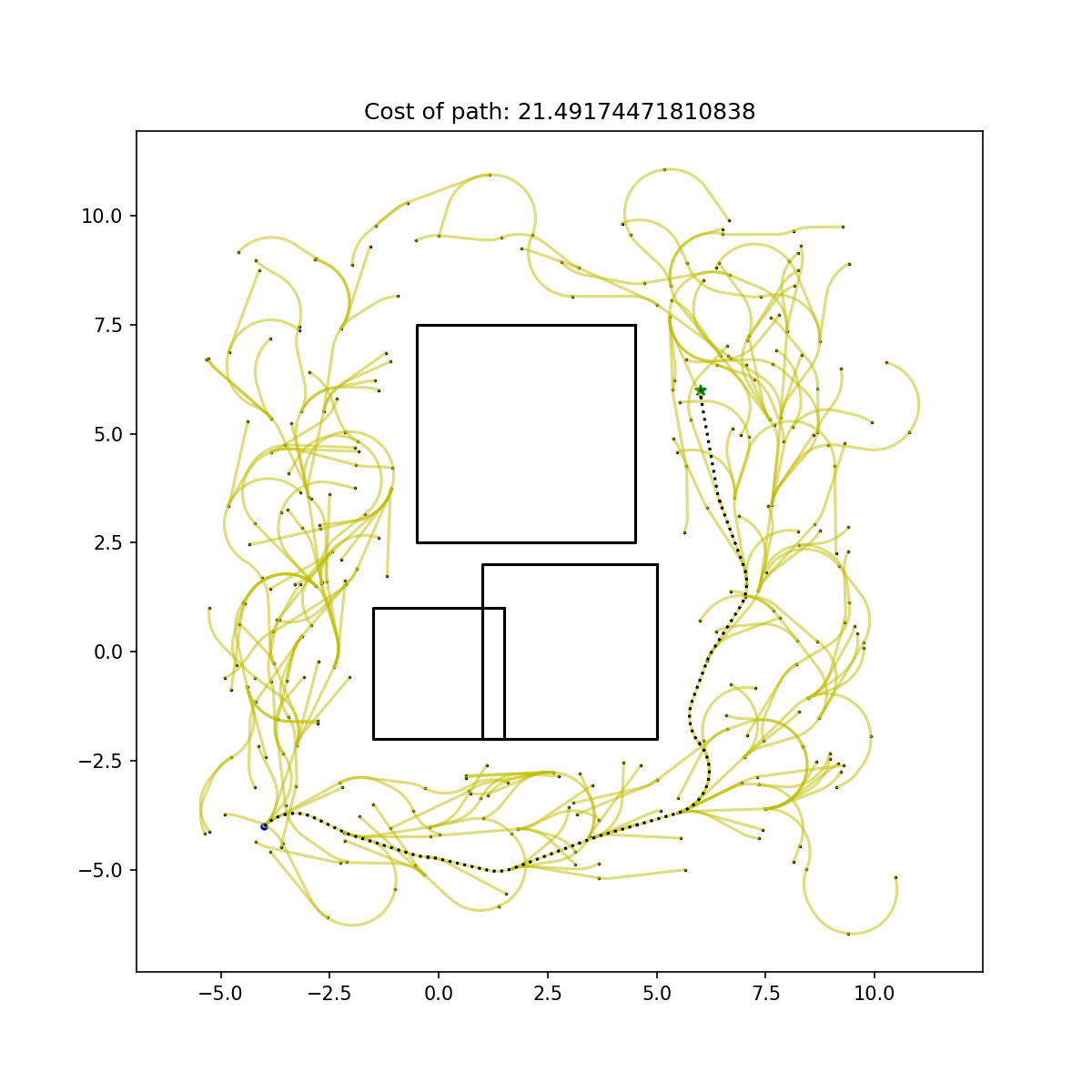}
  \caption{RRT}
  \label{fig:rrt_ass2}
\end{subfigure}%
\begin{subfigure}{.25\textwidth}
  \centering
  \includegraphics[width=\linewidth]{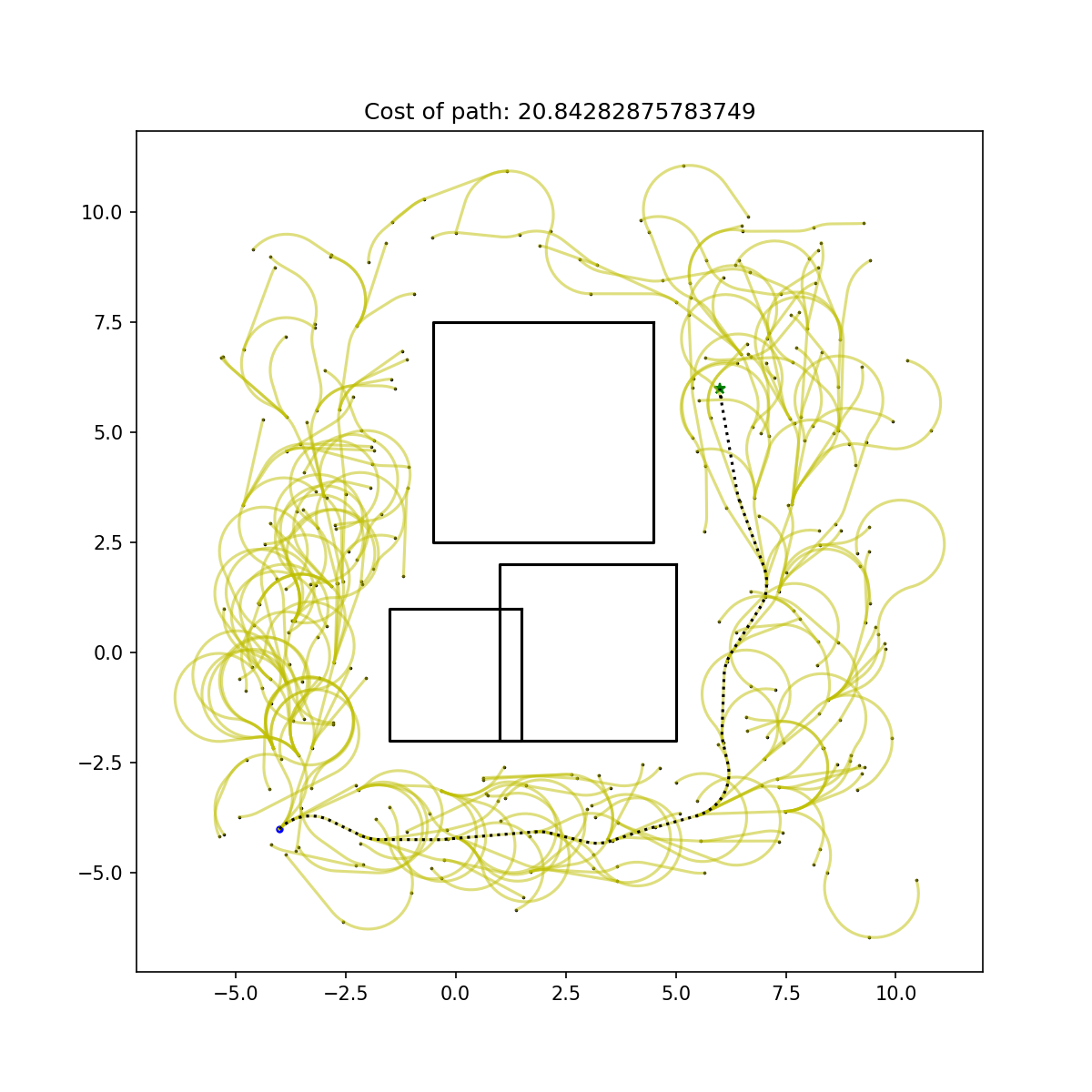}
  \caption{RRT*}
  \label{fig:rrtstar_ass2}
\end{subfigure}
\caption{RRT and RRT* for block environment}
\label{fig:ass2}
\end{figure}

\begin{figure}[h]
\centering
\begin{subfigure}{.25\textwidth}
  \centering
  \includegraphics[width=\linewidth]{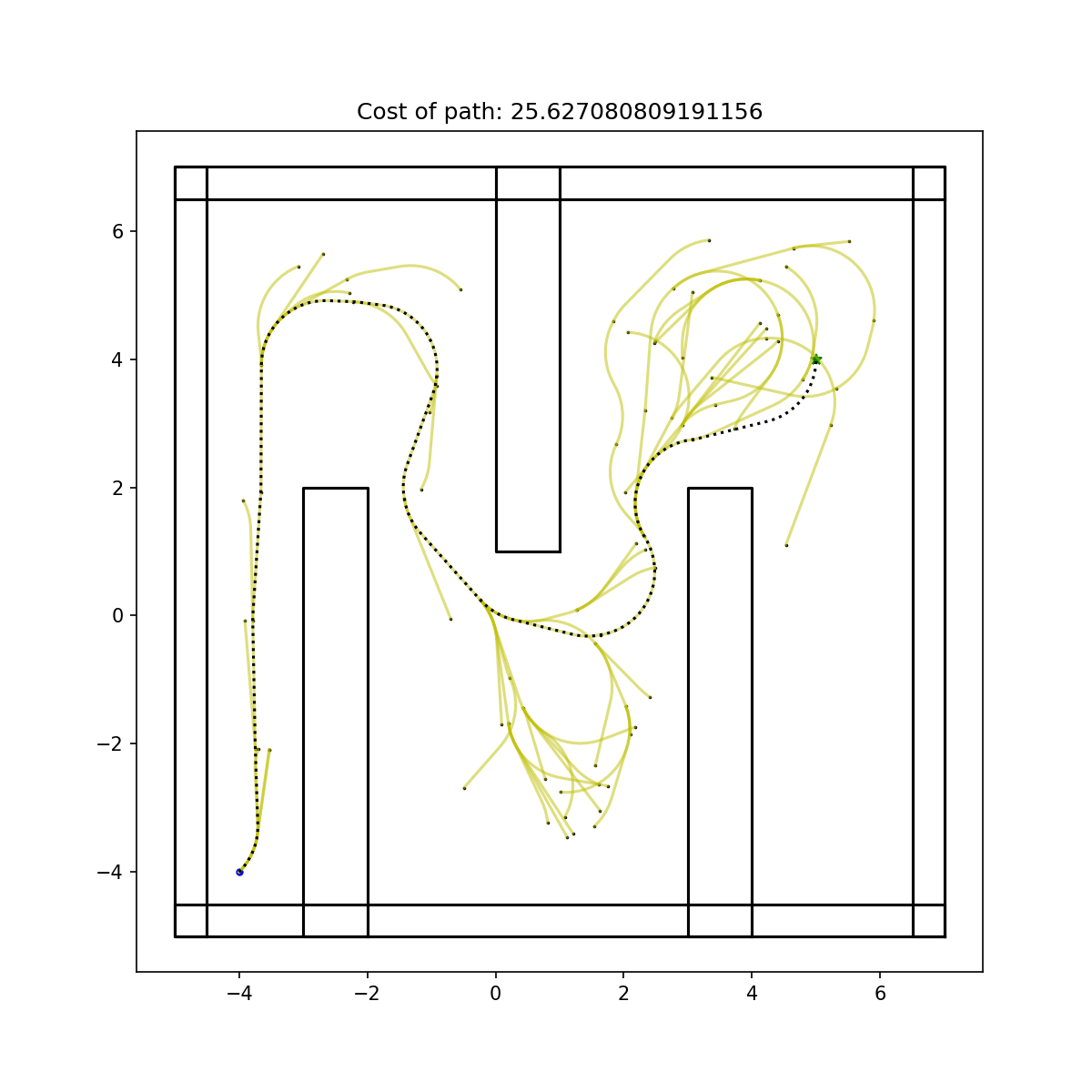}
  \caption{RRT}
  \label{fig:rrt_maze}
\end{subfigure}%
\begin{subfigure}{.25\textwidth}
  \centering
  \includegraphics[width=\linewidth]{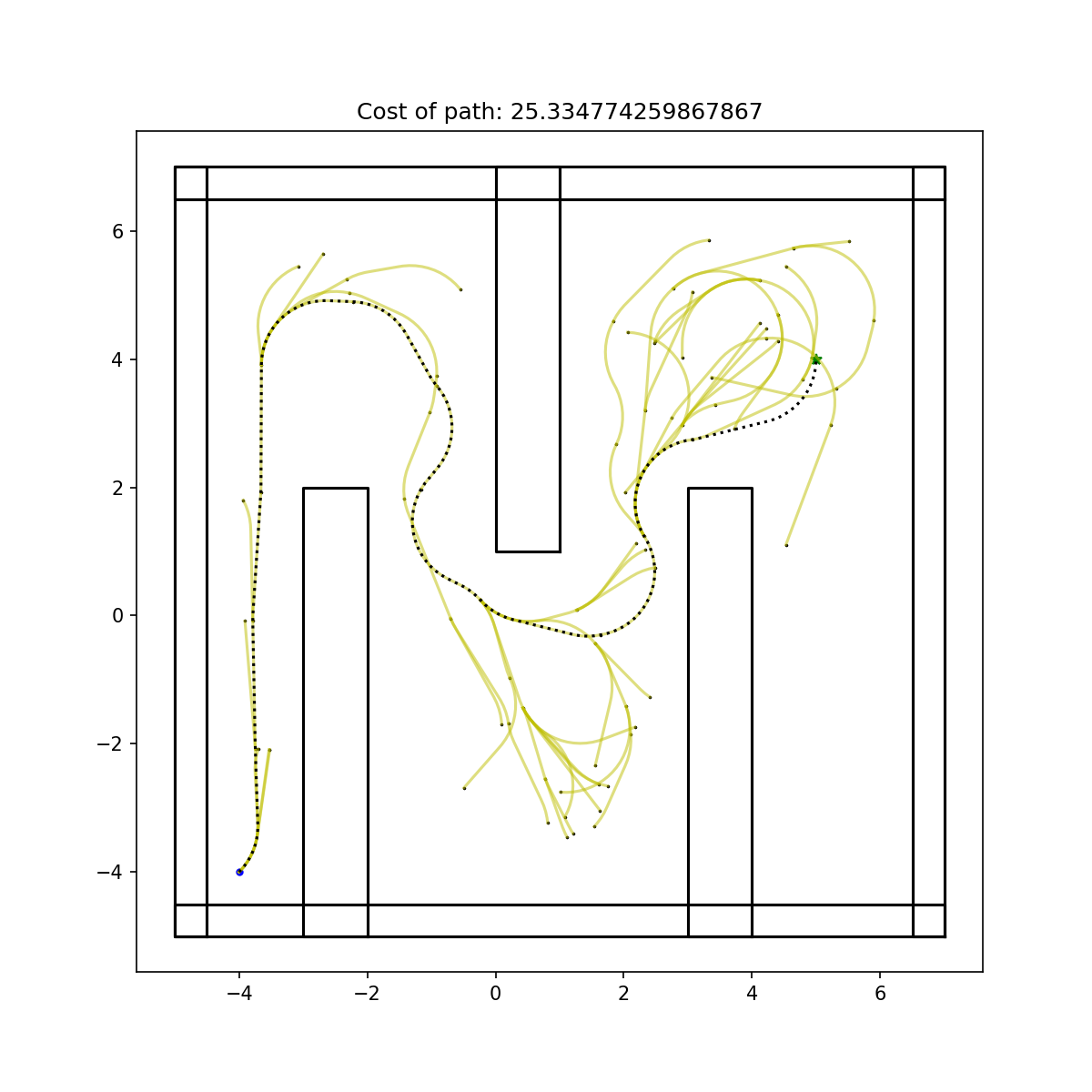}
  \caption{RRT*}
  \label{fig:rrtstar_maze}
\end{subfigure}
\caption{RRT and RRT* for maze environment}
\label{fig:maze}
\end{figure}

\begin{figure}[h]
\centering
\begin{subfigure}{.25\textwidth}
  \centering
  \includegraphics[width=\linewidth]{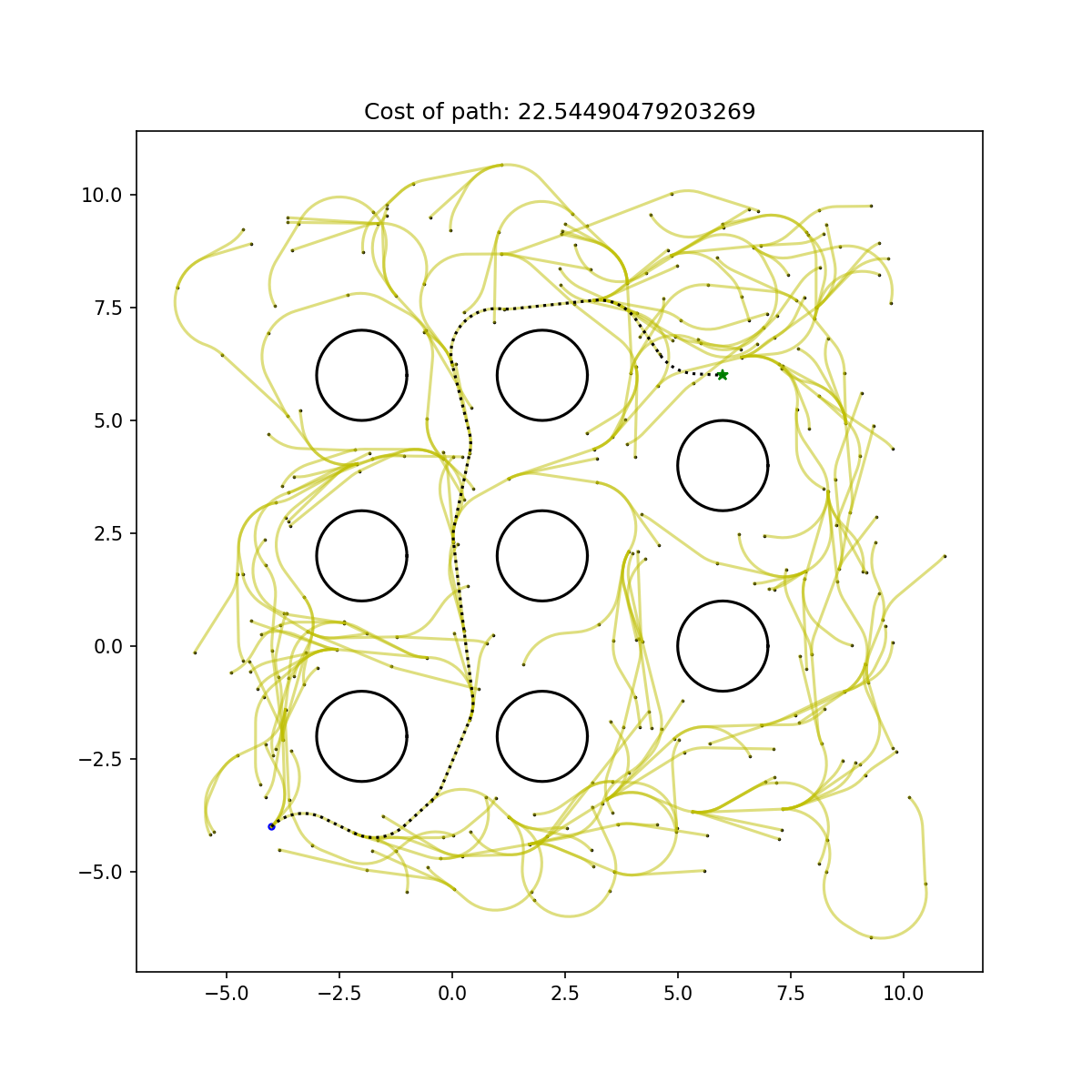}
  \caption{RRT}
  \label{fig:rrt_forest}
\end{subfigure}%
\begin{subfigure}{.25\textwidth}
  \centering
  \includegraphics[width=\linewidth]{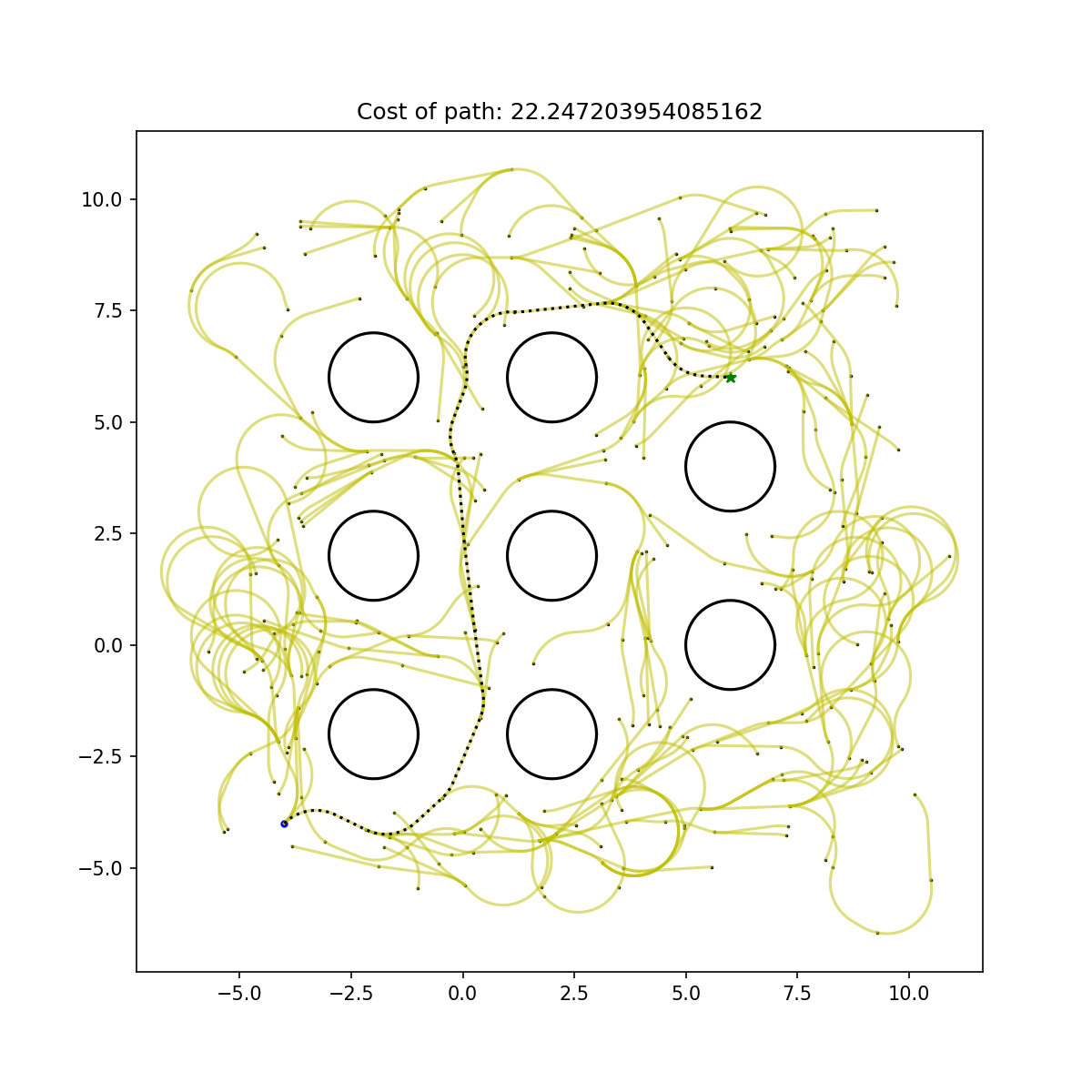}
  \caption{RRT*}
  \label{fig:rrtstar_forest}
\end{subfigure}
\caption{RRT and RRT* for forest environment}
\label{fig:forest}
\end{figure}

The modified algorithm for RRT and RRT* is simulated and tested in three different environment. Python package shapely \citep{shapely2007} is used to create obstacle in the grid. After the environment is created, the grid is inflated according to the vehicle shape to avoid collision when planning. For the vehicle dynamics configuration, the minimum turning radius is taken as $r_{min} = 1$ imposed by the maximum steering angle. 500 different points are sampled for RRT and RRT* in the environment to find the goal location.
The result for RRT and RRT* in block Environment, Maze Environment, Forest Environment is shown in fig \ref{fig:ass2},\ref{fig:maze},\ref{fig:forest}. We can see that, compared to RRT, the links are changed in RRT*. The overall cost for RRT* is lower compared to RRT. 

\begin{figure}[h]
\centering
\begin{subfigure}{.25\textwidth}
  \centering
  \includegraphics[width=\linewidth]{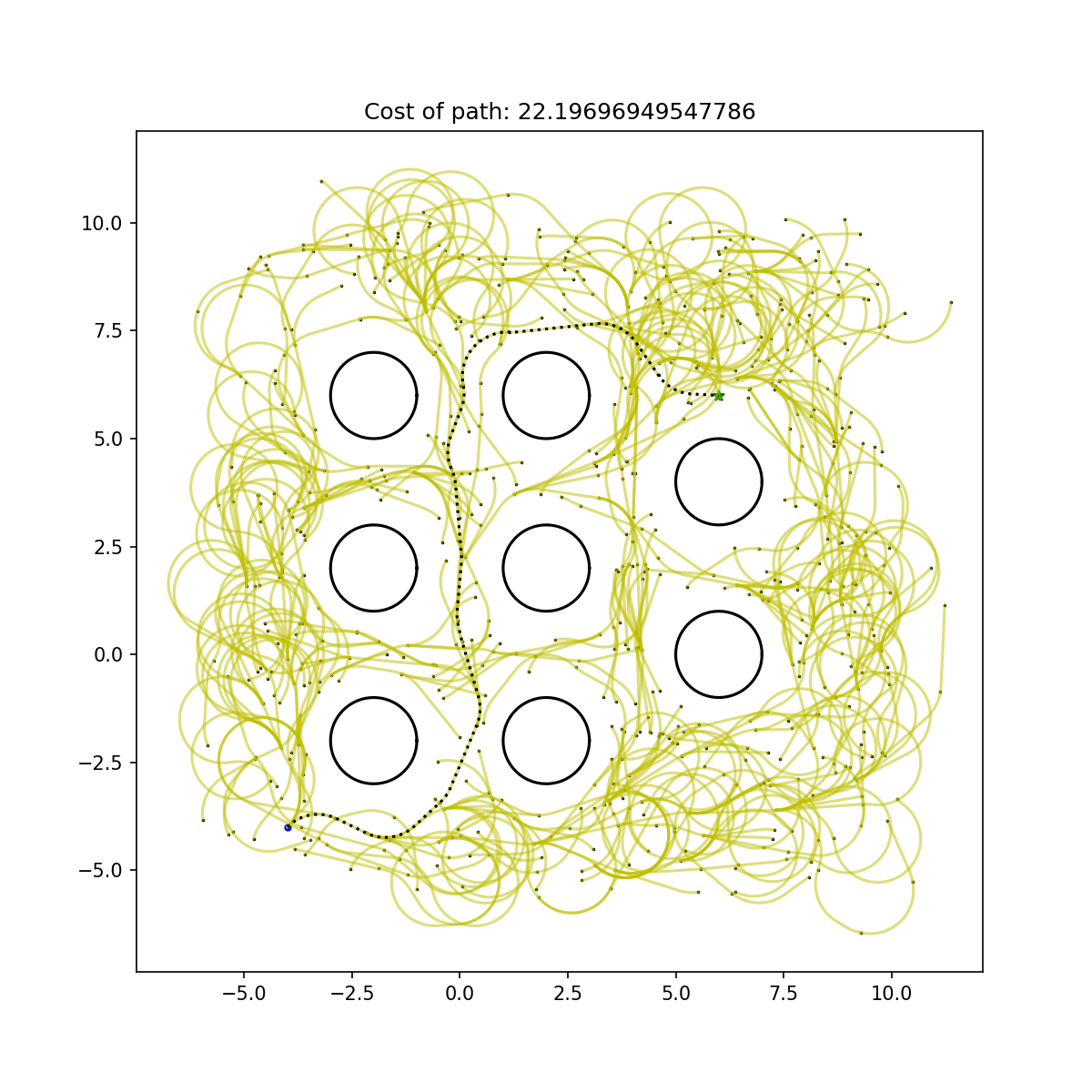}
  \caption{1000 sample points}
  \label{fig:rrtstar_1000}
\end{subfigure}%
\begin{subfigure}{.25\textwidth}
  \centering
  \includegraphics[width=\linewidth]{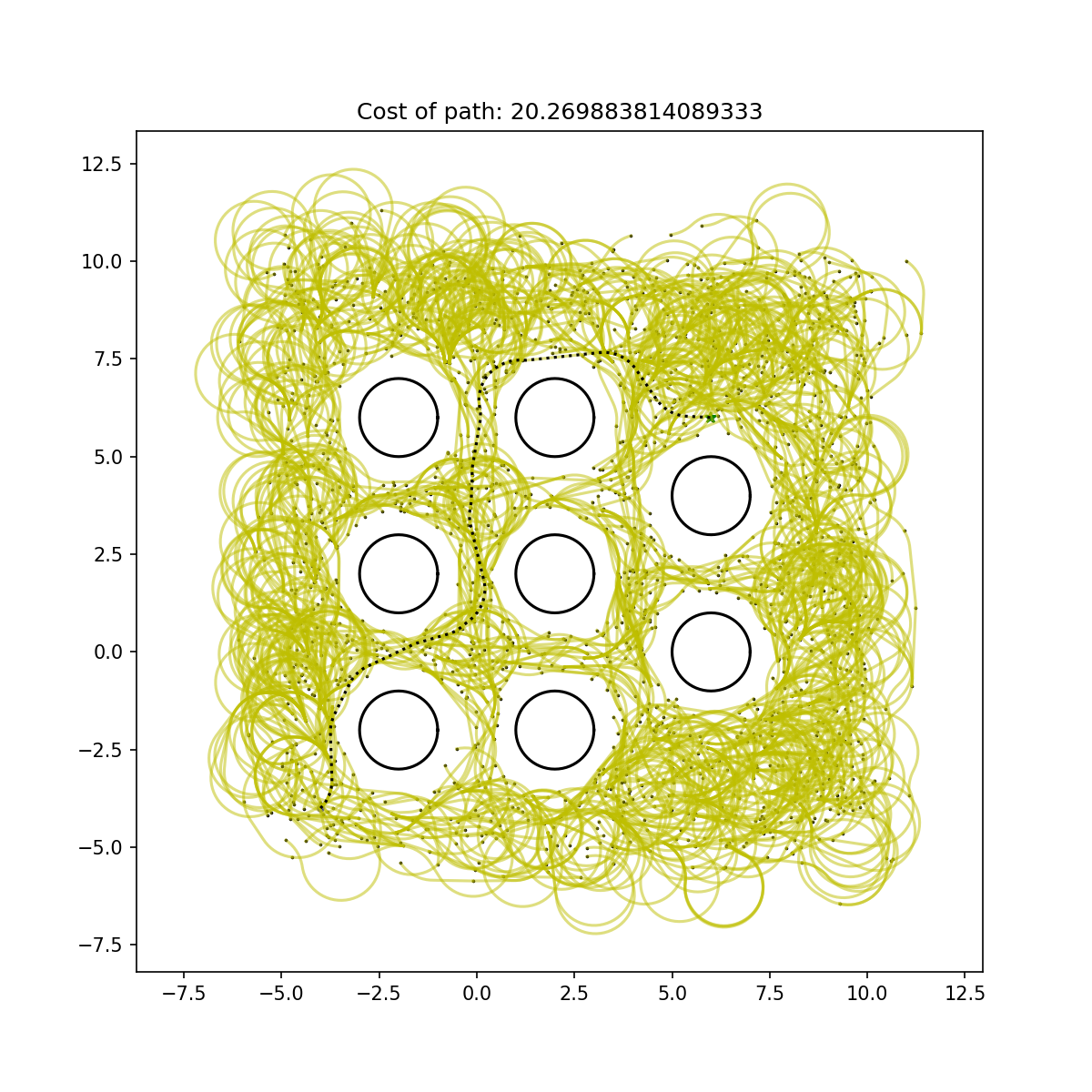}
  \caption{3000 sample points}
  \label{fig:rrtstar_3000}
\end{subfigure}
\caption{RRT* for different sample points}
\label{fig:rrtstar_iter}
\end{figure}

\begin{table}[!ht]
\centering
\begin{tabular}{|c|c|c|}
\hline
\textbf{\begin{tabular}[c]{@{}c@{}}sample\\ points\end{tabular}} & \textbf{\begin{tabular}[c]{@{}c@{}}RRT\\ Path length\end{tabular}} & \textbf{\begin{tabular}[c]{@{}c@{}}RRT*\\ Path length\end{tabular}} \\ \hline
500                                                                                                                            & 22.54                   & 22.24                                            \\ \hline
1000                                                                                                                         & 22.54                      & 22.19                                            \\ \hline
2000                                                                                                                         & 22.54                        & 20.75                                          \\ \hline
3000                                                                                                                       & 22.54                                & 20.26                                    \\ \hline

\end{tabular}
\caption{RRT and RRT* for different sample points}
\label{RRTvsRRTstar}
\end{table}

Modified RRT and RRT* algorithm are compared for different number of sample points in the maze environment. The result is shown in table \ref{RRTvsRRTstar}.

\section*{Discussion and Conclusion}
In this project, RRT and RRT* algorithm are implemented using vehicle dynamics. The position of random point is selected from uniform random sampling of the 2D space of the environment whereas the orientation (heading) of the random point is selected from uniform random sampling from $[0,2\pi]$. The results shown in table \ref{RRTvsRRTstar} shows less distance for RRT* compared to RRT. If more and more points are sampled, RRT* converges to optimality \citep{optimality}. An interesting observation would be when there is no limit to step size and the heading of the new random point always pointing towards goal. 

\bibliography{sections/references.bib}

\begin{thebibliography}{6}
\providecommand{\natexlab}[1]{#1}
\providecommand{\url}[1]{\texttt{#1}}
\expandafter\ifx\csname urlstyle\endcsname\relax
  \providecommand{\doi}[1]{doi: #1}\else
  \providecommand{\doi}{doi: \begingroup \urlstyle{rm}\Url}\fi

\bibitem[Dubins(1957)]{10.2307/2372560}
L.~E. Dubins.
\newblock On curves of minimal length with a constraint on average curvature,
  and with prescribed initial and terminal positions and tangents.
\newblock \emph{American Journal of Mathematics}, 79\penalty0 (3):\penalty0
  497--516, 1957.
\newblock ISSN 00029327, 10806377.
\newblock URL \url{http://www.jstor.org/stable/2372560}.

\bibitem[Gillies et~al.(2007)]{shapely2007}
S.~Gillies et~al.
\newblock Shapely: manipulation and analysis of geometric objects, 2007.
\newblock URL \url{https://github.com/Toblerity/Shapely}.

\bibitem[LaValle(1998)]{LaValle1998RapidlyexploringRT}
S.~M. LaValle.
\newblock Rapidly-exploring random trees : a new tool for path planning.
\newblock \emph{The annual research report}, 1998.

\bibitem[LaValle(2006)]{10.5555/1213331}
S.~M. LaValle.
\newblock \emph{Planning Algorithms}.
\newblock Cambridge University Press, USA, 2006.
\newblock ISBN 0521862051.

\bibitem[Shkel and Lumelsky(2001)]{SHKEL2001179}
A.~M. Shkel and V.~Lumelsky.
\newblock Classification of the dubins set.
\newblock \emph{Robotics and Autonomous Systems}, 34\penalty0 (4):\penalty0
  179--202, 2001.
\newblock ISSN 0921-8890.
\newblock \doi{https://doi.org/10.1016/S0921-8890(00)00127-5}.
\newblock URL
  \url{https://www.sciencedirect.com/science/article/pii/S0921889000001275}.

\bibitem[Solovey et~al.(2019)Solovey, Janson, Schmerling, Frazzoli, and
  Pavone]{optimality}
K.~Solovey, L.~Janson, E.~Schmerling, E.~Frazzoli, and M.~Pavone.
\newblock Revisiting the asymptotic optimality of rrt$^*$, 2019.
\newblock URL \url{https://arxiv.org/abs/1909.09688}.

\end{thebibliography}

\end{document}